\documentclass[10pt,twocolumn,letterpaper]{article}

\pdfoutput=1

\usepackage{cvpr}
\usepackage{times}
\usepackage{epsfig}
\usepackage{graphicx}
\usepackage{amsmath}
\usepackage{amssymb}

\usepackage{epstopdf}

\cvprfinalcopy 

\def\cvprPaperID{****} 

\begin{document}

\title{Deep AutoEncoder-based Lossy Geometry Compression for Point Clouds  }

\author{Wei Yan$^{1}$, Yiting Shao$^{2}$, Shan Liu$^{3}$, Thomas H Li$^{4}$, Zhu Li$^{5}$, Ge Li$^{6,\ast}$\\
$ ^{1,2,6}$School of Electronic and Computer Engineering, Peking University Shenzhen Graduate School\\
$ ^{3}$Tencent America\\
$ ^{4}$Advanced Institute of Information Technology, Peking University\\
$ ^{5}$School of Electronic and Computer Engineering, Peking University Shenzhen Graduate School\\
{\tt\small $ ^{1}$yanwe@pku.edu.cn, $^{6,\ast}$Corresponding Author:geli@ece.pku.edu.cn}
}

\maketitle

\begin{abstract}
   Point cloud is a fundamental 3D representation which is widely used in real world applications such as autonomous driving. As a newly-developed media format which is characterized by complexity and irregularity, point cloud creates a need for compression algorithms which are more flexible than existing codecs. Recently, autoencoders(AEs) have shown their effectiveness in many visual analysis tasks as well as image compression, which inspires us to employ it in point cloud compression.  In this paper, we propose a general autoencoder-based architecture for lossy geometry point cloud compression. To the best of our knowledge, it is the first autoencoder-based geometry compression codec that directly takes point clouds as input rather than voxel grids or collections of images. Compared with handcrafted codecs, this approach adapts much more quickly to previously unseen media contents and media formats, meanwhile achieving competitive performance. Our architecture consists of a pointnet-based encoder, a uniform quantizer, an entropy estimation block and a nonlinear synthesis transformation module. In lossy geometry compression of point cloud, results show that the proposed method outperforms the test model for categories 1 and 3 (TMC13) published by MPEG-3DG group on the 125th meeting, and on average a 73.15\% BD-rate gain is achieved.
   
\end{abstract}


\maketitle
\begin{figure*}
    \centering
    \includegraphics[width=1\textwidth,height=8cm]{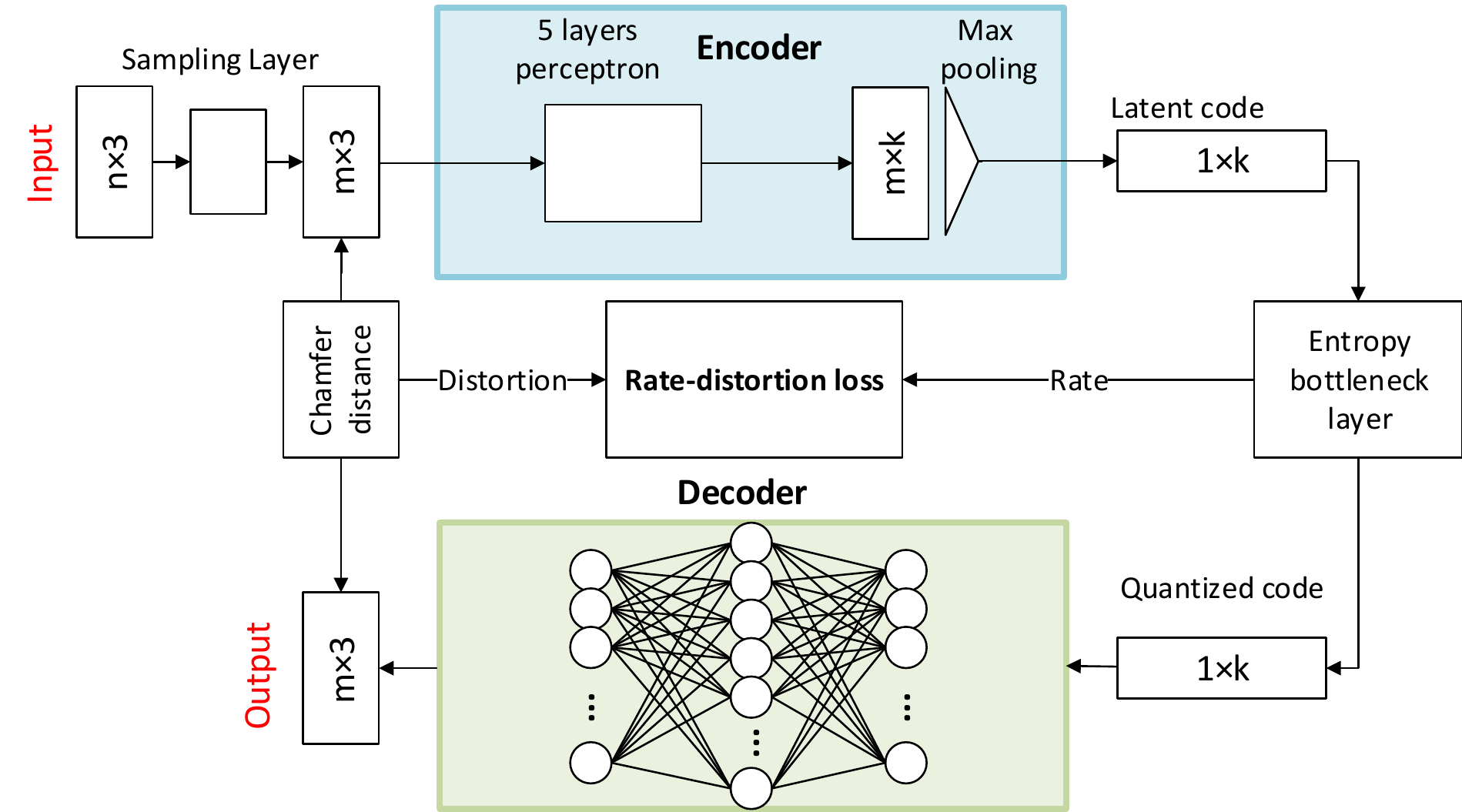}
    \caption{Compression architecture.}
    \label{fig:architecture}
\end{figure*}

\section{Introduction}
    Thanks to recent developments in 3D sensoring technology, point cloud has become a useful representation of holograms which enables free viewpoint viewing. It has been used in many fields such as Virtual/Augmented/Mixed reality (VR/AR/MR), smart city, robotics and automated driving\cite{mpeg}. 
 Point cloud compression becomes an increasingly important technique in order to efficiently process and transmit this type of data. It has attracted much attention from researchers as well as the MPEG Point Cloud Compression (PCC) group\cite{mpeg}.
 Geometry compression and attribute compression are two fundamental problems of static point cloud compression. Geometry compression aims at compressing the point locations. Attribute compression targets at reducing the redundancy among points' attribute values, given the point locations. This paper focuses on the geometry  compression.
     
 Point cloud is a set of unordered points that are irregularly distributed in Euclidean space. Because there is no uniform grid structure like 2D pictures, traditional image compression and video compression schemes cannot effectively work. In recent years, many researchers have been dedicated to developing methods for it. Octrees \cite{Jackins1980Oct}\cite{Meagher1982Geometric}\cite{Schnabel2006Octree} are usually used to compress geometry information of point clouds, and it has been developed for intra- and inter- frame coding. In the MPEG PCC group, point cloud compression is divided into two profiles: a video coding based method named V-PCC and a geometry based method named G-PCC.  All these methods are carefully designed by human experts who apply various heuristics to reduce the amount of information needing to be preserved and  to transform the resulting code in a way that is amenable to lossless compression. However, when designing a point cloud codec, human experts usually focus on a specific type of point cloud and tend to make assumptions about their features because of the diversity of point clouds. For example, an early version of the G-PCC reference software, TMC13, was divided into two parts, one for compressing point clouds belonging to category 1, and the other for compressing point clouds belonging to category 3, which means that it was difficult to build a universal point cloud codec. Therefore, given a particular type of point cloud, designing a codec that adapts quickly to the characteristics of such a point cloud and achieves better compression efficiency is a problem worth exploring.
 
In recent years, 3D machine learning has made great progress in high-level vision tasks such as classification and detection\cite{Charles2017PointNet}\cite{NIPS2017_7095}\cite{Zhou_2018_CVPR}. A natural question is whether we can employ this useful class of methods to further develop the point cloud codec, especially for point cloud sizes for which we do not have carefully designed. Usually, the design of a new point cloud codec can take years, but a compression framework based on neural networks may be able to adapt much more quickly to those niche tasks.
    
In this work, we consider the point cloud compression as an analysis/synthesis problem with a bottleneck in the middle. There are a  number of research aims to teach neural networks to discover compressive representations. Achlioptas et al.\cite{achlioptas2018learning} proposed an end-to-end deep auto-encoder that directly takes point clouds as inputs. Yang et al.\cite{Yang_2018_CVPR} further proposed a graph-based encoder and a folding-based decoder. These auto-encoders are able to extract compressive representations from point clouds in terms of the transfer classification accuracy. These works motivate us to develop a novel autoencoder-based lossy geometry point cloud codec. Our proposed architecture consists of four modules: a pointnet-based encoder, a uniform quantizer, an entropy estimation block and a nonlinear synthesis transformation module. To the best of our knowledge, it is the first autoencoder-based geometry compression codec that directly takes point clouds as input rather than voxel grids or collections of images.

\section{Related works}
Several approaches for point cloud geometry compression have been proposed in the literature.
Sparse Voxel Octrees (SVOs), also called Octrees\cite{Jackins1980Oct}\cite{Meagher1982Geometric}, are usually used to compress geometry information of point clouds\cite{Schnabel2006Octree}\cite{Huang2008A}\cite{8296514}\cite{Mekuria2016Design}\cite{Kammerl2012Real}. R. Schnabel et al.\cite{Schnabel2006Octree} first used Octrees in point cloud compression. This work predicts occupancy codes by surface approximations and uses octree structure to encode color information. Y. Huang et al. \cite{Huang2008A} further developed it for progressive point cloud coding. They reduced the entropy by bit reordering in the subdivision bytes. Their method also includes attribute coding such as color coding based on frequency of occurrence and normal coding using spherical quantization. There are several methods that adopt inter- and intra-frame coding in point cloud geometry compression\cite{8296514}\cite{motioncom}. Kammerl et al.\cite{Kammerl2012Real} developed a prediction octree and used XOR as an inter-coding tool. Mekuria et al\cite{Mekuria2016Design} further proposed an octree-based intra and inter coding system.
 In MPEG PCC\cite{mpeg}, the depth of the octree is constrained at a certain level and the leaf node of the octree is regarded as a single point, a list of points or a geometric model. Triangulations, also called triangle soups, are regarded as the geometric model in PCC. Pavez et al.\cite{Pavez2017Dynamic} first explored the polygon soup representation of geometry for point cloud compression.
 
 Recently, 3D machine learning has made great progress. Deep networks that directly handle points in a point set are state-of-the-art for supervised learning tasks on point clouds such as classification and segmentation. Qi et al.\cite{Charles2017PointNet}\cite{NIPS2017_7095} first proposed a deep neural network that directly takes point clouds as input. After that, many other networks were proposed for high-level analysis problems with point clouds\cite{Klokov2017Escape}\cite{Hua2017Point}\cite{pointcnn}\cite{Wang2017SGPN}. There are also a few works that focus on 3D autoencoders. Achlioptas et al.\cite{achlioptas2018learning} proposed an end-to-end deep auto-encoder that directly takes point clouds as input. Yang et al.\cite{Yang_2018_CVPR} further proposed a graph-based encoder and a folding-based decoder. To the best of our knowledge, there are few machine-learning-based works focusing on point cloud compression,
but several autoencoder-based methods have been proposed to enhance the performance of image compression. Toderici et al.\cite{Toderici2015Variable} proposed to use recurrent neural networks (RNNs) for image compression. Theis et al.\cite{Theis2017LossyIC} achieved multiple bit rates by learning a scaling parameter that changes the effective quantization coarseness. Ball$\acute{e}$ et al\cite{Ball2017end}\cite{Ball2018Variational}\cite{Ball2016End}. used a similar autoencoder architecture and replaced the non-differentiable quantization function with a continuous relaxation by adding uniform noise. 
\section{Formulation of point cloud geometry compression}
We consider a point cloud codec generally as an analysis/synthesis problem with a bottleneck in the middle. The input point cloud is represented as a set of 3D points $\{P_{i}|i=1,...,n\}$, where each point $P_i$ is a vector of its $(x,y,z)$ coordinate. In order to compress the input point cloud, the encoder should transform the input point cloud in Euclidean space $\mathbf{R^{3}}$ into higher dimensional feature space $\mathcal{H}$. In the feature space, we could discard some tiny components by quantization, which reduces the redundancy in the information. So we get the compressive representations as the latent code $\mathbf{z}$. Then, the decoder transforms the compressive representations from the feature space $\mathcal{H}$ back into Euclidean space $\mathbf{R^{3}}$ and we get the reconstructed point set $\{P_{i}^{'}|i=1,...,n\}$.

\section{Proposed geometry compression architecture}
In this section, we describe the proposed compression architecture. The details of each component will be discussed in the subsections. Our proposed architecture consists of four modules: a pointnet-based encoder, a uniform quantizer, an entropy estimation block and a nonlinear synthesis transformation module.
 We adopt auto-encoder\cite{Krizhevsky2011UsingVD} as our basic compression platform. The structure of the auto-encoder is shown in Figure \ref{fig:architecture}. 
 Firstly, the input point cloud is downsampled by the sampling layer $S$ to create a point cloud with different point density. Then, the downsampled point set goes through the autoencoder-based codec.
 The codec consists of an encoder $E$ that takes an unordered point set as input and produces a compressive representation, a quantizer $Q$, and a decoder $D$ that takes the quantized representation produced by $Q$ and produces a reconstructed point cloud. Thus, our compression architecture can be formulated as:
\begin{equation}
    x^{'}=D(Q(E(S(x)))),
\end{equation}
where $x$ is the original unordered point set and $x^{'}$ is the reconstructed point cloud.


\subsection{Sampling layer}
In G-PCC, the octree-based geometry coding uses a quantization scale to control the lossy geometry compression\cite{mpeg}. Let $(X_{i}=(x_{i},y_{i},z_{i}))_{i=1...N}$ be the set of 3D positions associated with the points of the input point cloud. The G-PCC encoder computes the quantized positions $(\hat{X}_{i})_{i=1...N}$ as follows:
\begin{equation}
    \hat{X}_{i}=\lfloor (X_{i}-X_{shift})\times s \rfloor,
\end{equation}
where $X_{shift}$ and $s$ are user-defined parameters that are signaled in the bitstream. After quantization, there will be many duplicate points sharing the same quantized positions. A common approach is merging those duplicate points, which reduces the number of points in the input point cloud.

Inspired by G-PCC, we use a sampling layer\cite{NIPS2017_7095} to achieve the \textit{downsampling} step. Given input points $\{x_{1},x_{2},...,x_{n}\}$, we adopt iterative farthest point sampling (FPS)\cite{NIPS2017_7095} to select a subset of points $\{x_{i_{1}},x_{i_{2}},...,x_{i_{m}}\}$, such that $x_{i_{j}}$ is the farthest point (in metric distance) from the former point set $\{x_{i_{1}},x_{i_{2}},...,x_{i_{j-1}}\}$ with regard to the rest points. In contrast to random sampling, the point density of the resulted point set is more uniform, which is better at keeping the shape characteristics of the original object.


\subsection{Encoder and Decoder}
Generally, an autoencoder can be regarded as an analysis function, $y=f_e(x;\theta_{e})$, and a synthesis function, $\hat{x}=f_{d}(y;\phi_{d})$,where $x$, $\hat{x}$, and $y$ are original point clouds, reconstructed point clouds, and compressed data, respectively.        $\theta_e$ and $\phi_d$ are optimized parameters in the analysis and the synthesis function.

To learn the encoded compressive representation, we consider the pointnet architecture\cite{Charles2017PointNet}\cite{achlioptas2018learning}. There are $n$ points in the input point set ($n\times3$ matrix). Every point is encoded by several 1-D convolution layers with kernel size 1. Each convolution layer is followed by a ReLU\cite{Nair2010Rectified} and a batch-normalization layer\cite{Ioffe2015Batch}. In order to make a model invariant to input permutation, there is a feature-wise maximum layer following the last convolutional layer to produce a $k$-dimensional latent code. This latent code will be quantized and the quantized latent code will be encoded by the entropy encoder to get the final bitstream. In our experiment, we use 5 1-D convolutional layers and the number of filters in each layer is 64, 128, 128, 256 and k respectively. k is decided by the number of input points. See details in the experimental results.

Currently, there are two kinds of decoder for point clouds: the fully-connected decoder\cite{achlioptas2018learning} and the folding-based decoder\cite{Yang_2018_CVPR}. Both decoders are able to produce reconstructed point clouds. The folding-based decoder is much smaller in parameter size than the fully-connected decoder, but there will be more hyperparameters to choose such as the number of grid points and the interval of grid points. Thus, we choose the fully-connected decoder as our decoder, which interprets the latent code using 3 fully-connected layers to produce a $n\times3$ reconstructed point cloud. Each fully-connected layer is followed by a ReLU, and the number of nodes in each hidden layer is 256, 256 and $n\times3$ respectively. 


\subsection{Quantization}

To reduce the amount of information necessarily needed for storage and transmission, quantization is a significant step in media compression. However, quantization functions like the rounding function's derivative is zero or undefined. Theis et al.\cite{Theis2017LossyIC}  replaced the derivative in the backward pass of backpropagation with the derivative of a smooth approximation $r$. Thus, the rounding function's derivative becomes:
\begin{equation}
    \frac{d}{dz}[z]=\frac{d}{dz}r(z).
    \label{smooth}
\end{equation}
During backpropagation, the derivative of the rounding function will be computed by equation (\ref{smooth}), while the rounding function will not be replaced by the smooth approximation during the forward pass\cite{Theis2017LossyIC}. This is because if we  replace the rounding function with the smooth approximation completely, the decoder may learn to invert the smooth approximation, thereby affecting the entropy bottleneck layer that learns the entropy model of the latent code. In \cite{Theis2017LossyIC}, they found $r(z)=z$ to work as well as more sophisticated choices.

In contrast to \cite{Theis2017LossyIC}, Ball$\acute{e}$ et al.\cite{Ball2016End} replaced quantization by additive uniform noise,
\begin{equation}
    [f(x)] \approx f(x)+u,
\end{equation}
where $u$ is random noise. 
 After experiment, we find that the performance of these two methods is very similar. In our implementation, we use the second method.
 
 \begin{figure*}
    \centering
    \includegraphics[width=1.02\textwidth]{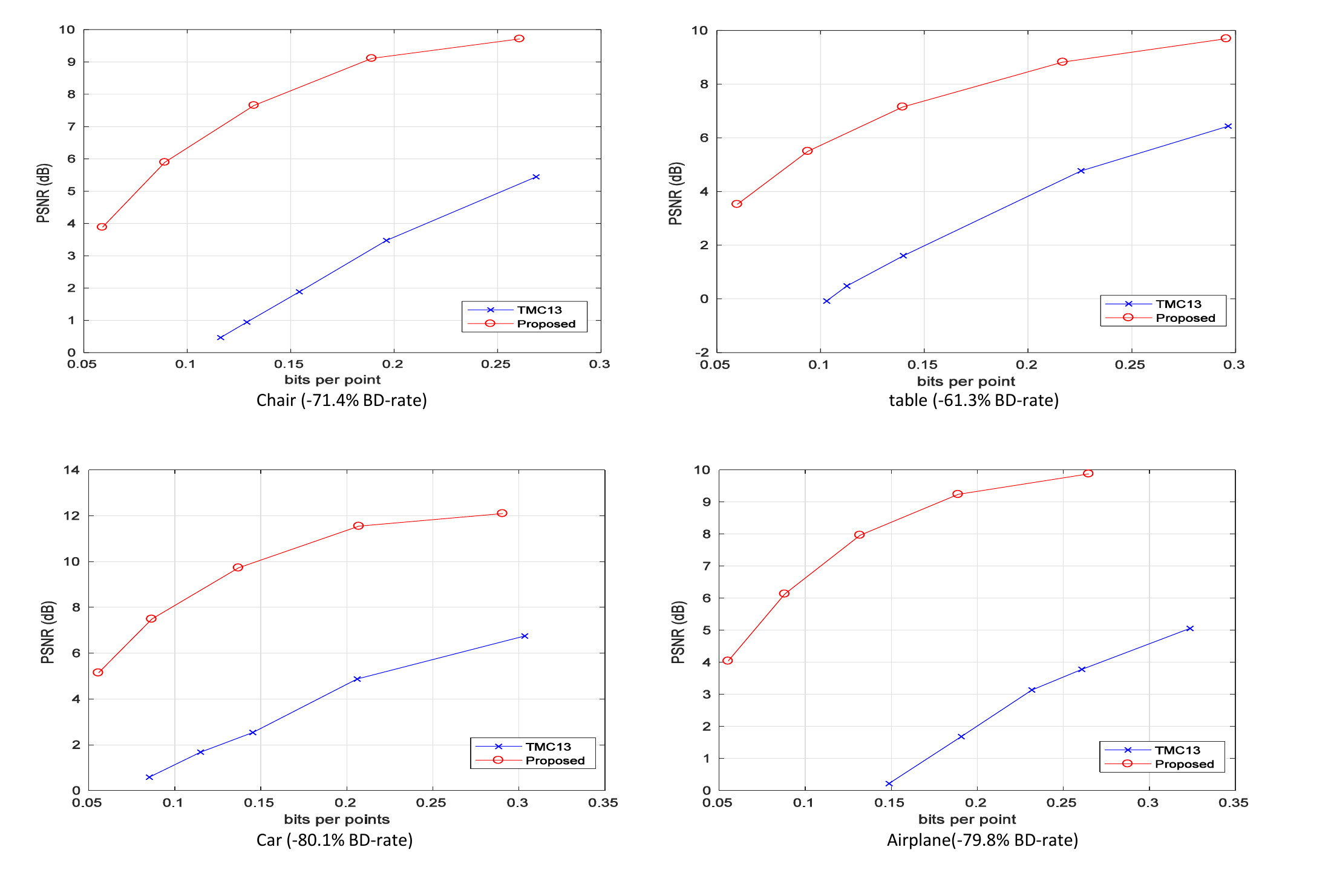}
    \caption{Objective results.}
    \label{fig:objective}
\end{figure*}

\begin{figure*}
    \centering
    \includegraphics[width=0.91\textwidth,height=0.98\textheight]{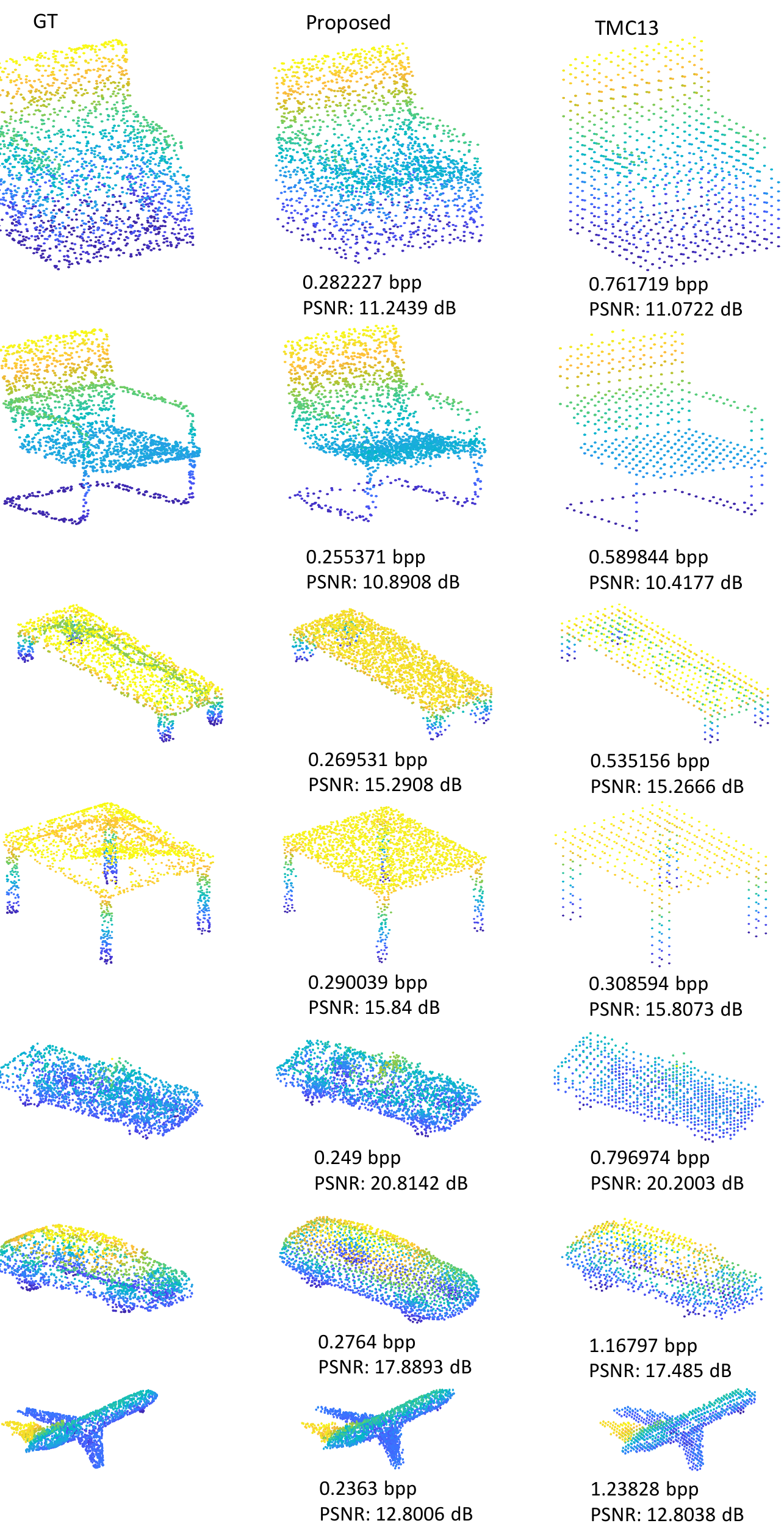}
    \caption{Subjective results. Left: Original point clouds (GT). Color for better display.}
    \label{fig:subjective}
\end{figure*}

\begin{figure}
    \centering
    \includegraphics[width=0.52\textwidth,height=0.4\textwidth]{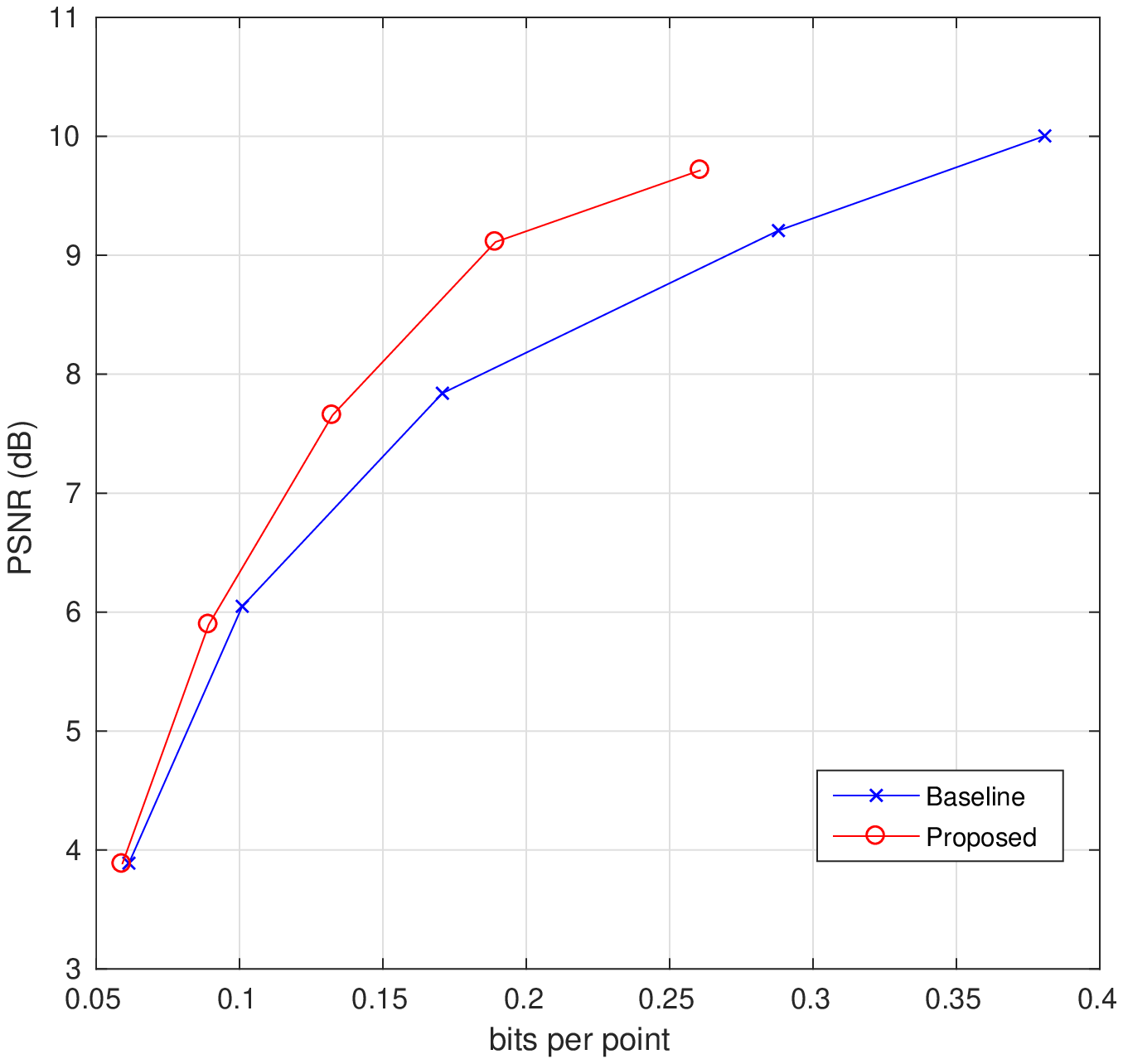}
    \caption{An ablation study of entropy estimation module. The results are tested on point clouds of chair class. A 19.3\% BD-rate gain is obtained.}
    \label{fig:ablation}
\end{figure}

\subsection{Rate-distortion Loss}
In the lossy compression problem, one must trade off between the entropy of the discretized latent code ($rate$) and the error caused by the compression ($distortion$). So, our goal is to minimize the weighted sum of the rate (R) and the distortion (D) $\lambda D+R$ over the parameters of encoder, decoder and the rate estimation model that will be discussed later, where $\lambda$ controls the tradeoff. 

Entropy rate estimation has been studied by many researchers who use neural networks to compress images\cite{Theis2017LossyIC} \cite{Ball2017end} \cite{Ball2018Variational}. In our architecture, the encoder $E$ transforms the input point cloud $x$ into a latent representation $z$, using a non-linear function $f_{e}(x;\theta_e)$ with parameters $\theta_e$. The latent code $z$ is then quantized to form $\hat{z}$ by quantizer $Q$. $\hat{z}$ can be losslessly compressed by entropy coding techniques such as arithmetic coding as its value is discrete. The rate of the discrete code $\hat{z}$, $R$, is lower-bounded by the entropy of the discrete probability distribution of $\hat{z}$, $H[m_{\hat{z}}]$, that is:
\begin{equation}
    R_{min} = \mathbb{E}_{\hat{z} \sim{m}}[-\log_{2}m_{\hat{z}}(\hat{z})],
\end{equation}
where $m(\hat{z})$ is the actual marginal distribution of the discrete latent code. However, the $m(\hat{z})$ is unknown to us and we need to estimate it by building a probability model according to some prior knowledge. Suppose we get an estimation of the probability model $p_{\hat{z}}(\hat{z})$. Then the actual rate is given by the Shannon cross entropy between the marginal $m(\hat{z})$ and the prior $p_{\hat{z}}(\hat{z})$:
\begin{equation}
    R=\mathbb{E}_{\hat{z} \sim{m}} [-\log_{2}p_{\hat{z}}(\hat{z})].
\end{equation}
Therefore, if the estimated model distribution is identical to the actual marginal distribution, the rate is minimum and the estimated rate is the most accurate. Similar to \cite{Ball2018Variational}, we use the entropy bottleneck layer\footnote{\url{https://tensorflow.github.io/compression/docs/entropy_bottleneck.html}} that models the prior  $p_{\hat{z}}(\hat{z})$ using a non-parametric and fully factorized density model:
\begin{equation}
    p_{\hat{z}|\psi }(\hat{z}|\psi)=\prod_{i} \left( p_{z_{i} |\psi^{(i)}}(\psi^{(i)})*u(-\frac{1}{2},\frac{1}{2})   \right)(\hat{z}_{i}),
\end{equation}
where the vector $\psi^{(i)}$ represents the parameters of each univariate distribution $p_{z_{i}|\psi^{(i)}}$ 
. Note that each non-parametric density is convolved with a standard uniform density, which enables a better match of the prior to the marginal\cite{Ball2018Variational}. 

The distortion is computed by the Chamfer distance. Suppose that there are $n$ points in the original point cloud, represented by a $n\times3$ matrix. Each row of the matrix is composed of the 3D position $(x,y,z)$. The reconstructed point cloud is represented by a $m\times3$ matrix. The number of original points $n$ may be different from $m$ because of the lossy compression. Suppose the original point cloud is $S_1$ and the reconstructed point set is $S_2$. Then, the reconstruction error is computed by the Chamfer distance:
\begin{equation}
    d_{CH}(S_{1},S_{2})=\sum_{x\in S_1} \min_{y\in S_2} \lVert x-y\rVert_{2}^{2}+\sum_{y\in S_2} \min_{x\in S_1}\lVert x-y\rVert_{2}^{2}.
\end{equation}

Finally, our rate-distortion loss function is:
\begin{equation}
    L[f_{e},f_{d},p_{\hat{z}}]=\lambda \mathbb{E}[d_{CH}(S1,S2)]+\mathbb{E}[-\log_{2}p_{\hat{z}}],
\end{equation}
where $f_e$ is the non-linear function of the encoder, $f_d$ is the non-linear function of the decoder and $p_{\hat{z}}$ is the estimated probability model. The expectation will be approximated by averages over a training set of point clouds.


\section{Experimental results}

\subsection{Datasets}
Since our data-driven method requires a large number of point clouds for training, we use the ShapeNet dataset\cite{shapenet2015}. Shapes from ShapeNet dataset are axis aligned and centered into the unit sphere. The  point-cloud-format of the ShapeNet data set is obtained by uniformly sampling points on the triangles from the mesh models in the dataset. Without additional statements, we train models with point clouds from a single object class and the train/test splits is 90\%-10\%.

\subsection{Implementation Details}
The number of points in the original point cloud is 2048 and the latent code's dimension is 512. To better compare with the reconstructed point clouds that compressed by TMC13, we downsample the original point cloud to 1024, 512, 256 and 128 points and the corresponding latent code sizes are 256, 128, 64 and 32. Because the TMC13 can not compress the normalized point cloud directly, we expand normalized point clouds to be large enough so that the TMC13 can compress it properly. To be fair, we also perform the same operation when we compress point clouds by our model. We add an extra normalization before feeding point clouds into our network, and expand reconstructed point clouds when we compute their distortion.

We implement our model based on the python2.7 platform with Tensorflow 1.12. We run our model in a computer with an i7-8700 CPU and a GTX1070 GPU (with 8G memory). We use the Adam\cite{DBLP:journals/corr/KingmaB14} method to train our network with learning rate 0.0005. We train our model with entropy optimization within 1200 epochs and train the model without entropy optimization within 500 epochs. The batch size is 8 as limited by our GPU memory.

\subsection{Compression results}

We compare our method with the TMC13 anchor latest released in the MPEG 125th meeting.
We experiment on four types of point clouds: chair, airplane, table and car. These four types of point clouds contain a very rich point cloud shape. The chair category contains 6101 point clouds for training and 678 point clouds for testing. The airplane category contains 3640 point clouds for training and 405 point clouds for testing. The table category contains 7658 point clouds for training and 851 point clouds for testing. The car category contains 6747 point clouds for training and 750 point clouds for testing. 
Rate-distortion performances for each types of point cloud  are shown in Figure \ref{fig:objective}. To avoid unfairly penalizing the TMC13 due to the unavoidable cost of file headers, we exclude the header size from bitstream produced by TMC13. The distortion in Figure \ref{fig:objective} is the point-to-point geometry PSNR obtained from the \textit{pc\_error} MPEG tool\cite{pcerror}. Rate-distortion curves are obtained by averaging over all test point clouds. 
Results show that our method outperforms the TMC13 in all types of point clouds at all bitrates. On average, a 73.15\% BD-rate gain can be achieved.

In Figure \ref{fig:subjective} we show some test point clouds compressed to low bit rates. In line with objective results, we find that our method produces smaller bits per point than TMC13 under the similar PSNR reconstruction quality. The reconstructed point clouds of proposed method is more dense than those compressed by TMC13. 

To further analyze the entropy estimation module in our method, we implement a simple ablation study. We consider the model without the entropy bottleneck layer as our baseline. The comparison of the RD curve between the baseline and our proposed model on point clouds of chair is presented in Figure \ref{fig:ablation}. Results show that the entropy estimation can effectively reduce the size of bitstream, yielding a 19.3\% BD-rate gain.



\section{Conclusion}
In this paper, we propose a general deep autoencoder-based architecture for lossy geometry point cloud compression.
Compared with handcrafted codecs, this approach not only achieves better coding efficiency, but also can adapt much quicker to new media contents and new media formats. Experimental evaluation demonstrates that for the given benchmark, the proposed model outperforms the TMC13 on the rate-distortion performance, and on average a 73.15\% BD-rate gain is achieved.


To the best of our knowledge, it is the first autoencoder-based geometry compression codec that directly takes point clouds as input rather than voxel grids or collections of images. The algorithms that we present may also be extended to work on attribute compression of point cloud or even point cloud sequence compression. To encourage future work, we will make all the materials public.


{\small
\bibliographystyle{ieee}
\bibliography{sample_base}

\begin{thebibliography}{10}\itemsep=-1pt

\bibitem{achlioptas2018learning}
P.~Achlioptas, O.~Diamanti, I.~Mitliagkas, and L.~Guibas.
\newblock Learning representations and generative models for 3d point clouds,
  2018.

\bibitem{Ball2017end}
J.~Ballé, V.~Laparra, and E.~Simoncelli.
\newblock End-to-end optimized image compression.
\newblock 11 2016.

\bibitem{Ball2016End}
J.~Ballé, V.~Laparra, and E.~P. Simoncelli.
\newblock End-to-end optimization of nonlinear transform codes for perceptual
  quality.
\newblock In {\em Picture Coding Symposium}, 2016.

\bibitem{Ball2018Variational}
J.~Ballé, D.~Minnen, S.~Singh, S.~J. Hwang, and N.~Johnston.
\newblock Variational image compression with a scale hyperprior.
\newblock 2018.

\bibitem{shapenet2015}
A.~X. Chang, T.~Funkhouser, L.~Guibas, P.~Hanrahan, Q.~Huang, Z.~Li,
  S.~Savarese, M.~Savva, S.~Song, H.~Su, J.~Xiao, L.~Yi, and F.~Yu.
\newblock {ShapeNet: An Information-Rich 3D Model Repository}.
\newblock Technical Report arXiv:1512.03012 [cs.GR], Stanford University ---
  Princeton University --- Toyota Technological Institute at Chicago, 2015.

\bibitem{Charles2017PointNet}
R.~Q. Charles, S.~Hao, K.~Mo, and L.~J. Guibas.
\newblock Pointnet: Deep learning on point sets for 3d classification and
  segmentation.
\newblock In {\em IEEE Conference on Computer Vision \& Pattern Recognition},
  2017.

\bibitem{motioncom}
R.~L. {de Queiroz} and P.~A. {Chou}.
\newblock Motion-compensated compression of point cloud video.
\newblock In {\em 2017 IEEE International Conference on Image Processing
  (ICIP)}, pages 1417--1421, Sep. 2017.

\bibitem{8296514}
D.~C. {Garcia} and R.~L. {de Queiroz}.
\newblock Context-based octree coding for point-cloud video.
\newblock In {\em 2017 IEEE International Conference on Image Processing
  (ICIP)}, pages 1412--1416, Sep. 2017.

\bibitem{Hua2017Point}
B.~S. Hua, M.~K. Tran, and S.~K. Yeung.
\newblock Point-wise convolutional neural network.
\newblock 2017.

\bibitem{Huang2008A}
Y.~Huang, J.~Peng, C.~C. Kuo, and M.~Gopi.
\newblock A generic scheme for progressive point cloud coding.
\newblock {\em IEEE Transactions on Visualization \& Computer Graphics},
  14(2):440--453, 2008.

\bibitem{Ioffe2015Batch}
S.~Ioffe and C.~Szegedy.
\newblock Batch normalization: accelerating deep network training by reducing
  internal covariate shift.
\newblock In {\em International Conference on International Conference on
  Machine Learning}, 2015.

\bibitem{Jackins1980Oct}
C.~L. Jackins and S.~L. Tanimoto.
\newblock Oct-trees and their use in representing three-dimensional objects.
\newblock {\em Computer Graphics \& Image Processing}, 14(3):249--270, 1980.

\bibitem{Kammerl2012Real}
J.~Kammerl, N.~Blodow, R.~B. Rusu, M.~Beetz, E.~Steinbach, and S.~Gedikli.
\newblock Real-time compression of point cloud streams.
\newblock In {\em IEEE International Conference on Robotics \& Automation},
  2012.

\bibitem{DBLP:journals/corr/KingmaB14}
D.~P. Kingma and J.~Ba.
\newblock Adam: {A} method for stochastic optimization.
\newblock In {\em {ICLR}}, 2015.

\bibitem{Klokov2017Escape}
R.~Klokov and V.~Lempitsky.
\newblock Escape from cells: Deep kd-networks for the recognition of 3d point
  cloud models.
\newblock In {\em 2017 IEEE International Conference on Computer Vision
  (ICCV)}, 2017.

\bibitem{Krizhevsky2011UsingVD}
A.~Krizhevsky and G.~E. Hinton.
\newblock Using very deep autoencoders for content-based image retrieval.
\newblock In {\em ESANN}, 2011.

\bibitem{pointcnn}
Y.~Li, R.~Bu, M.~Sun, and B.~Chen.
\newblock Pointcnn.
\newblock 01 2018.

\bibitem{Meagher1982Geometric}
D.~Meagher.
\newblock Geometric modeling using octree encoding.
\newblock {\em Computer Graphics \& Image Processing}, 19(2):129--147, 1982.

\bibitem{Mekuria2016Design}
R.~Mekuria, K.~Blom, and P.~Cesar.
\newblock Design, implementation and evaluation of a point cloud codec for
  tele-immersive video.
\newblock {\em IEEE Transactions on Circuits \& Systems for Video Technology},
  PP(99):1--1, 2016.

\bibitem{Nair2010Rectified}
V.~Nair and G.~E. Hinton.
\newblock Rectified linear units improve restricted boltzmann machines.
\newblock In {\em International Conference on International Conference on
  Machine Learning}, 2010.

\bibitem{Pavez2017Dynamic}
E.~Pavez and P.~A. Chou.
\newblock Dynamic polygon cloud compression.
\newblock In {\em IEEE International Conference on Acoustics}, 2017.

\bibitem{NIPS2017_7095}
C.~R. Qi, L.~Yi, H.~Su, and L.~J. Guibas.
\newblock Pointnet++: Deep hierarchical feature learning on point sets in a
  metric space.
\newblock In I.~Guyon, U.~V. Luxburg, S.~Bengio, H.~Wallach, R.~Fergus,
  S.~Vishwanathan, and R.~Garnett, editors, {\em Advances in Neural Information
  Processing Systems 30}, pages 5099--5108. Curran Associates, Inc., 2017.

\bibitem{Schnabel2006Octree}
R.~Schnabel and R.~Klein.
\newblock Octree-based point-cloud compression.
\newblock In {\em Eurographics}, 2006.

\bibitem{mpeg}
S.~Schwarz, M.~Preda, V.~Baroncini, M.~Budagavi, P.~Cesar, P.~Chou, R.~Cohen,
  M.~Krivokuća, S.~Lasserre, Z.~Li, J.~Llach, K.~Mammou, R.~Mekuria,
  O.~Nakagami, E.~Siahaan, A.~Tabatabai, A.~M.~Tourapis, and V.~Zakharchenko.
\newblock Emerging mpeg standards for point cloud compression.
\newblock {\em IEEE Journal on Emerging and Selected Topics in Circuits and
  Systems}, PP:1--1, 12 2018.

\bibitem{Theis2017LossyIC}
L.~Theis, W.~Shi, A.~Cunningham, and F.~Husz{\'a}r.
\newblock Lossy image compression with compressive autoencoders.
\newblock {\em CoRR}, abs/1703.00395, 2017.

\bibitem{pcerror}
D.~Tian, H.~Ochimizu, C.~Feng, R.~Cohen, and A.~Vetro.
\newblock Geometric distortion metrics for point cloud compression.
\newblock pages 3460--3464, 09 2017.

\bibitem{Toderici2015Variable}
G.~Toderici, S.~M. O'Malley, S.~J. Hwang, D.~Vincent, D.~Minnen, S.~Baluja,
  M.~Covell, and R.~Sukthankar.
\newblock Variable rate image compression with recurrent neural networks.
\newblock {\em Computer Science}, 2015.

\bibitem{Wang2017SGPN}
W.~Wang, R.~Yu, Q.~Huang, and U.~Neumann.
\newblock Sgpn: Similarity group proposal network for 3d point cloud instance
  segmentation.
\newblock 2017.

\bibitem{Yang_2018_CVPR}
Y.~Yang, C.~Feng, Y.~Shen, and D.~Tian.
\newblock Foldingnet: Point cloud auto-encoder via deep grid deformation.
\newblock In {\em The IEEE Conference on Computer Vision and Pattern
  Recognition (CVPR)}, June 2018.

\bibitem{Zhou_2018_CVPR}
Y.~Zhou and O.~Tuzel.
\newblock Voxelnet: End-to-end learning for point cloud based 3d object
  detection.
\newblock In {\em The IEEE Conference on Computer Vision and Pattern
  Recognition (CVPR)}, June 2018.

\end{thebibliography}
}

\end{document}